\documentclass[10pt,twocolumn,letterpaper]{article}

\usepackage[pagenumbers]{cvpr} 

\usepackage{graphicx}
\usepackage{amsmath}
\usepackage{amssymb}
\usepackage{bm}
\usepackage{booktabs}
\usepackage{float}
\usepackage{cuted}
\usepackage{capt-of}
\usepackage{overpic}
\usepackage{multirow}
\usepackage{lipsum}
\usepackage{rotating}
\usepackage{gensymb}

\usepackage[pagebackref,breaklinks,colorlinks]{hyperref}

\usepackage[capitalize]{cleveref}
\crefname{section}{Sec.}{Secs.}
\Crefname{section}{Section}{Sections}
\Crefname{table}{Table}{Tables}
\crefname{table}{Tab.}{Tabs.}

\begin{document}

\title{SelfNeRF: Fast Training NeRF for Human from Monocular Self-rotating Video}
\author{\large Bo Peng  \quad Jun Hu \quad Jingtao Zhou\quad Juyong Zhang\thanks{Corresponding author} \vspace{0.5 mm}\\
{\normalsize University of Science and Technology of China}\\
{\tt\footnotesize \{ pb15881461858@mail.,hu997372@mail.,ustc$\_$zjt@mail.,juyong@\}ustc.edu.cn} \\}



\twocolumn[{
\maketitle

\vspace*{-8mm}

\begin{center}
   \begin{overpic}
        [width=\linewidth]{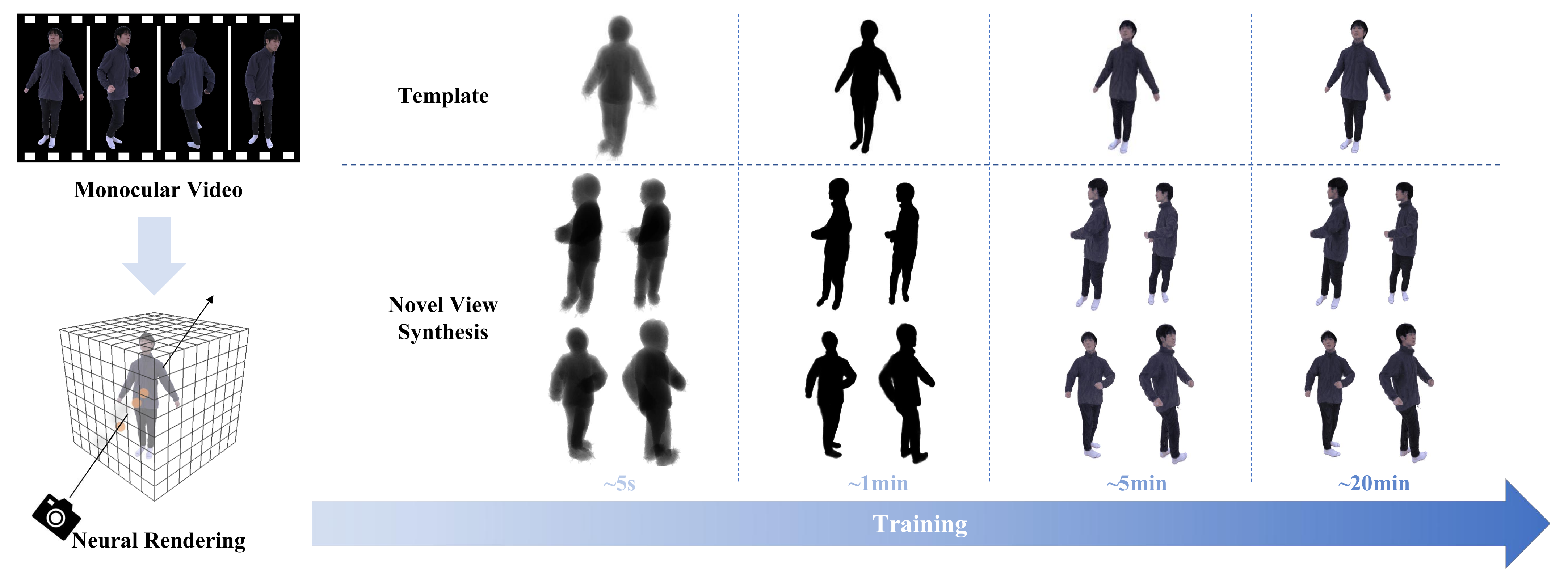}

   \end{overpic}
\end{center}
\vspace*{-5mm}
\captionof{figure}{Given a monocular self-rotating video of the human performer, SelfNeRF is able to train from scratch and converge in about twenty minutes, and then generate free-view points videos.}
\label{fig:teaser}

\vspace*{5mm}

}]

{
  \renewcommand{\thefootnote}%
    {\fnsymbol{footnote}}
  \footnotetext[1]{Corresponding Author}
}

\begin{abstract}
In this paper, we propose SelfNeRF, an efficient neural radiance field based novel view synthesis method for human performance. Given monocular self-rotating videos of human performers,  SelfNeRF can train from scratch and achieve high-fidelity results in about twenty minutes. Some recent works have utilized the neural radiance field for dynamic human reconstruction. However, most of these methods need multi-view inputs and require hours of training, making it still difficult for practical use. To address this challenging problem, we introduce a surface-relative representation based on multi-resolution hash encoding that can greatly improve the training speed and aggregate inter-frame information.
Extensive experimental results on several different datasets demonstrate the effectiveness and efficiency of SelfNeRF to challenging monocular videos. Our code and video results will be available at \href{https://ustc3dv.github.io/SelfNeRF}{https://ustc3dv.github.io/SelfNeRF}.

\end{abstract}


\section{Introduction}
\label{sec:intro}
Novel view synthesis of human performance is an important research problem in computer vision and computer graphics, and has wide applications in many areas such as sports event broadcasts, video conferences, and VR/AR. 
Although this problem has been widely studied for a long time, existing methods still require multi-camera systems and quite a long computation time.
These shortcomings cause this technology not easily used by public users.
Therefore, a high-fidelity novel view synthesis of human performance based on a monocular camera and training within tens of minutes will have significant value for practical use.

Traditional novel view synthesis methods need dense inputs for 2D image-based methods \cite{hedman2018deep} or require depth cameras for high-fidelity 3D reconstruction \cite{dou2016fusion4d} to render realistic results.
Some model-based methods\cite{Bogo:ECCV:2016,alldieck2018video,kolotouros2019convolutional} could reconstruct explicit 3D mesh from sparse RGB videos, but lack geometry detail and tend to be unrealistic. 
Recently, several works have applied NeRF\cite{mildenhall2020nerf} to synthesize novel view images of dynamic human bodies. NeuralBody\cite{peng2021neural}, AnimatableNeRF\cite{peng2022animatable}, H-NeRF\cite{xu2021h} and other works\cite{su2021nerf,zhao2021humannerf,kwon2021neural,liu2021neural} are able to synthesize high-quality rendering images and extract rough body geometry from sparse-view videos of the human body by combining human body priors with the NeRF model. 
However, most of these works require quite a long time to train for each subject. 
HumanNeRF\cite{zhao2021humannerf} does not require training for each subject from scratch but still takes around an hour to fine-tune the model to achieve better results, making it still difficult to put into practical use. 
The long training time of these methods is caused by the expensive computation cost of NeRF. 
Moreover, most of these works still need calibrated multi-view camera system to integrate multi-frame information to produce a consistent registration sequence, making it hard to deploy. 
Recently, with the well-designed multi-resolution hash encoding\cite{muller2022instant}, the training speed of NeRF has been improved by several orders. 
However, the current strategy of INGP\cite{muller2022instant} only works for static scenes with multi-view inputs, and how to extend it to dynamic scenes with monocular inputs has not yet been explored.

In this paper, we propose SelfNeRF, a view synthesis method for human body, which can synthesize high-fidelity novel view results of human performance with a monocular camera and can converge within tens of minutes. 
These characteristics make SelfNeRF practical for ordinary users.  
We achieve these targets via a novel surface-relative hash encoding by extending multi-resolution hash encoding\cite{muller2022instant} to dynamic objects while aggregating information across frames. 
Specifically, given the monocular self-rotation video of a human performer, we recover the surface shape for each frame with existing reconstruction methods like VideoAvatar\cite{alldieck2018video} and SelfRecon\cite{jiang2022selfrecon}. 
We then calculate the K-nearest neighbor points and signed distance on the current frame's point cloud for each query point and take the corresponding point of the k-nearest neighbor points on the canonical space and signed distance as relative representation. 
For a sample point at a specific frame, we first calculate its' relative representation, and then use hash encoding to get a high-dimensional feature fed to NeRF MLP to regress its color and density. 
We adopt volume rendering\cite{lombardi2019neural} to get the color of each pixel and then train our model with photometric loss and geometric guidance loss. Extensive experimental results demonstrate the effectiveness of our proposed method. 
In summary, the contributions of this paper include the following aspects:
\begin{itemize}
\item To the best of our knowledge, SelfNeRF is the first work that applies hash encoding to dynamic objects, which can reconstruct a dynamic neural radiance field of a human in tens' of minutes.
\item A surface-relative hash encoding is proposed to aggregate inter-frame information and significantly speed up the training of the neural radiance field for humans.
\item With the state-of-the-art clothed human body reconstruction method, we can reconstruct high-fidelity novel view synthesis of human performance with a monocular camera.
\end{itemize}

\section{Related Work}
\paragraph{Neural Radiance Field based Human Reconstruction}
NeRF(neural radiance field)\cite{mildenhall2020nerf} represents a static scene as a learnable 5D function and adopts volume rendering to render the image from any given view direction.
Though vanilla NeRF only fits for static scenes, requires dense view inputs, and is slow to train and render, lots of work has been done to improve NeRF to dynamic scenes\cite{pumarola2021d} and sparse view inputs\cite{niemeyer2021regnerf} and increase the training and rendering speed\cite{garbin2021fastnerf}.   
Recently, some researchers have focused on applying the neural radiance field to human reconstruction. 
Neuralbody\cite{peng2021neural} utilizes a set of latent codes anchored to a deformable mesh which is shared at different frames. 
H-NeRF\cite{xu2021h} employs a structured implicit human body model to reconstruct the temporal motion of humans. 
AnimatableNeRF\cite{peng2022animatable} introduces deformation fields based on neural blend weight fields to generate observation-to-canonical correspondences. 
Surface-Aligned NeRF\cite{xu2022surface} defines the neural scene representation on the mesh surface points and signed distances from the surface of a human body mesh. 
Neural Actor\cite{liu2021neural} integrates texture map features to refine volume rendering. 
HumanNeRF\cite{zhao2021humannerf} employs an aggregated pixel-alignment feature and a pose embedded non-rigid deformation field for tackling dynamic motions. 
A-NeRF\cite{su2021nerf} proposes skeleton embedding serves as a common reference that links constraints across time. 
Neural Human Performer\cite{kwon2021neural} introduces a temporal transformer and a multi-view transformer to aggregate corresponding features across space and time.
Weng et al.\cite{weng2022humannerf} optimize for NeRF representation of the person in a canonical T-pose and a motion field that maps the estimated canonical representation to every frame of the video via backward warps, making it only requires monocular inputs.
AD-NeRF\cite{guo2021adnerf} employs a conditional NeRF to generate audio-driven talking head.
HeadNeRF\cite{hong2021headnerf} adds controllable codes to NeRF to obtain the parametric representation of the human head.
Although these methods can generate novel view synthesis results for human, they still need several views of videos or are costly to train and evaluate.

\paragraph{Acceleration of Neural Radiance Field Training}
Although NeRF\cite{mildenhall2020nerf} could generate high-fidelity novel view synthesis, its long training time cannot be accepted in practical use. 
Therefore, how to improve the training speed of NeRF has been widely studied since its emergence of NeRF.
DS-NeRF\cite{kangle2021dsnerf} utilizes the depth information supplied by 3D point clouds to speed up convergence and synthesize better results from fewer training views. 
KiloNeRF\cite{Reiser2021ICCV} adopts thousands of tiny MLPs instead of one single large MLP, which could achieve real-time rendering and can train 2\textasciitilde 3x faster.
Plenoxels\cite{yu2021plenoxels} represent a scene as a sparse 3D grid with spherical harmonics and thus can be optimized without any neural components. 
DVGO\cite{sun2021direct} adopts a representation consisting of a density voxel grid and a feature voxel grid with a network for view-dependent appearance. 
DIVeR\cite{wu2021diver} utilizes a similar scene representation but applies deterministic rather than stochastic estimates during volume rendering.
Recently, INGP\cite{muller2022instant} proposed to store the features of the voxel grid in a multi-resolution hash table and employ a spatial hash function to query the features, thus can significantly reduce the number of optimizable parameters.

\paragraph{Human Shape Reconstruction from Images}
Some traditional model-based works only require the single view RGB input. SMPLify\cite{Bogo:ECCV:2016} utilizes SMPL\cite{SMPL:2015} model to represent human body and obtain per-frame parameters via optimization. 
SMPL+D based method Videoavatars\cite{alldieck2018video} first calculates per-frame poses using the SMPL model, then optimizes for the subject's shape in the canonical T-pose. 
Kolotouros et al.\cite{kolotouros2019convolutional} adopt GraphCNN to directly regress the 3D location of the SMPL template mesh vertices, relaxing the heavy reliance on the model's parameter space. 
Though these methods cannot achieve photo-realistic view synthesis due to the limitation of the explicit parametric model, their fast generated human surface can introduce human priors for implicit methods.\par
Instead of optimizing parameters per scene, some works utilize networks to learn the priors of humans from ground truth data. 
PIFu\cite{saito2019pifu} concatenates pixel’s aligned feature and depth of a given query point as the input of an MLP to obtain a 3D human occupancy field.
PIFuHD\cite{saito2020pifuhd} adds normal information to improve the geometric details. 
StereoPIFu\cite{hong2021stereopifu} introduces the volume alignment feature and relative z-offset when giving a pair of stereo videos, which can effectively alleviate the depth ambiguity and restore absolute scale information.
BCNet\cite{jiang2020bcnet} propose a layered garment representation and can support more garment categories and recover more accurate geometry.\par
To capture better geometry surface of humans, representing the human body as the zero isosurface of an SDF(signed distance field) has become popular. 
PHORHUM\cite{alldieck2022photorealistic} modifies the geometric representation to SDF to get finer geometry and normal, so they can simultaneously estimate detailed 3D geometry and the unshaded surface color together with the scene illumination.
ICON\cite{xiu2021icon} infers a 3D clothed human meshes from a color image by utilizing a body-guided normal prediction model and a local-feature-based implicit 3D representation conditioned on SMPL(-X).
SelfRecon\cite{jiang2022selfrecon} represents the human body as a template mesh and SDF in canonical space and utilizes a deformation field consisting of rigid forward LBS deformation and small non-rigid deformation to generate correspondences. 
Given monocular self-rotation RGB inputs, these methods are capable of generating meshes of clothed humans.

\section{Method}

\begin{figure}[htb]
    \centering
    \includegraphics[width=\linewidth]{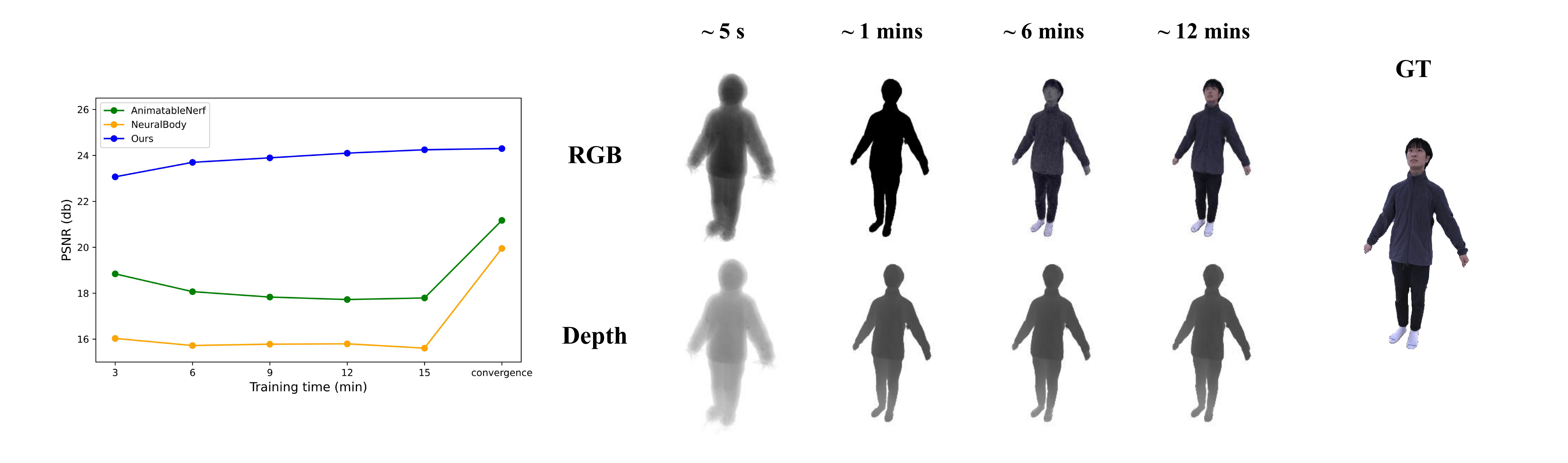}
    \vspace*{-7mm}
    \caption{Convergence speed of SelfNeRF. For a monocular videos with 200 frames, our method only requires twenty minutes to achieve reasonable results.}
    \label{fig:speed}
    \vspace*{-4mm}
\end{figure}

Given a monocular self-rotation video of a human performer, we aim to train from scratch and generate a free-viewpoint video of the performer in tens of minutes. 
For the input video, we recover the surface shape $\left\{{\mathbf{T}}_{t} \mid t=0, \ldots, N_{t}-1\right\}$ 
and the body mask $\left\{\mathbf{M}_{t} \mid t=0 \ldots, N_{t}-1\right\}$, where $t$ is the index of frame, $N_{t}$ is the total number of frames.

\begin{figure*}
    \centering
    \includegraphics[width=\textwidth]{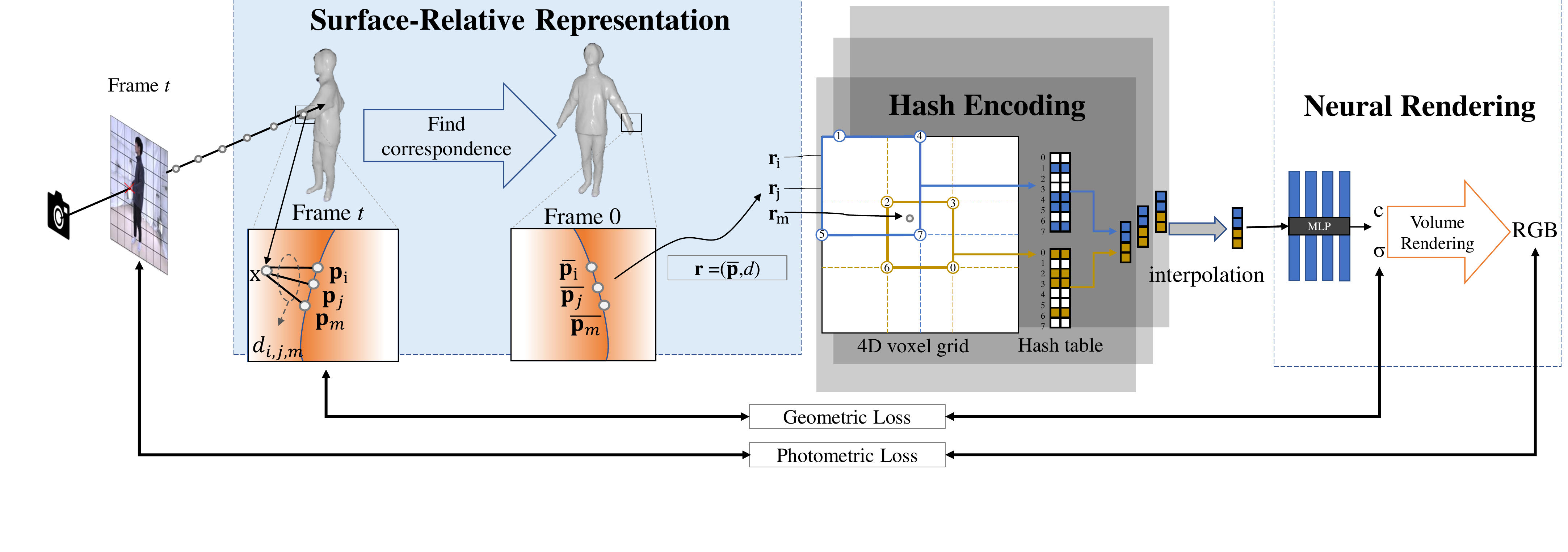}
    \vspace*{-13.5mm}
    \caption{Overview of SelfNeRF. Given sample point $\mathbf{x}$ at frame $t$, we first obtain a surface-relative representation $\mathbf{r}_i$ conditioned on human body surface via KNN of $\mathbf{p}_i$ to aggregate the corresponding point information of different frames. Then we exploit multi-resolution hash encoding to get the feature, which is the encoded input to the NeRF MLP to regress color and density at $\mathbf{x}$.}
    \label{fig:pipeline}
    \vspace*{-4.0mm}
\end{figure*}

Fig.~\ref{fig:pipeline} shows the overview of SelfNeRF. We first provide some background in Sec\ref{background}. 
Then in Sec\ref{representation}, we describe our dynamic human representation, in which a surface-relative hash encoding (see Sec\ref{relative}) is proposed to aggregate inter-frame information and significantly speed up training. 
Finally, we discuss our training strategy and loss in  sec\ref{loss}.
\subsection{Background}
\label{background}
\paragraph{Neural Radiance Field} A neural radiance field (NeRF)\cite{mildenhall2020nerf} represents a static 3D scene as a MLP function $F_{\omega}$ with learnable weights ${\omega}$ that outputs the radiance emitted in each direction $\mathbf{v}$ at each point $\mathbf{x}$ in space, and a density at each point. Then a volume rendering strategy is used to render the neural radiance field for any given camera pose.
In practice, for any query point $\mathbf{x}$ and view direction $\mathbf{v}$, NeRF encodes them with a positional encoding $\gamma$ that projects a coordinate vector into a high-dimensional space. 
These high dimensional vectors are then fed into $F_{\omega}$ to predict density $\sigma(\mathbf{x})$ and radiance $\mathbf{c}(\mathbf{x}, \mathbf{v})$ at input point $\mathbf{x}$. While rendering, NeRF samples one ray $\mathbf{r} = \mathbf{o} + u\mathbf{v}$ per pixel, and then calculates the pixel's color by the following volume rendering strategy\cite{lombardi2019neural}:
\begin{equation}
\centering
\mathbf{C}(\mathbf{r}) = \sum_{i = 1}^{N} \alpha\left(\mathbf{x}_{i}\right) \prod_{j<i}\left(1-\alpha\left(\mathbf{x}_{j}\right)\right) \mathbf{c}\left(\mathbf{x}_{i},\mathbf{v}\right),
\label{volume_rendering}
\end{equation}
where $\alpha\left(\mathbf{x}_{i}\right) = 1-\exp \left(-\sigma\left(\mathbf{x}_{i}\right) \delta_{i}\right)$, $\mathbf{x}_{i} =  \mathbf{o} + t_{i}\mathbf{v}$ are sample points on the ray, and $\delta_{i} = {u}_{i+1}-{u}_{i}$ are the distance between adjacent sample points. NeRF optimizes $F_{\omega}$ by photometric loss.

\paragraph{Multi-resolution Hash Encoding}
To increase the training speed, we adopt the multi-level hash encoding in INGP\cite{muller2022instant}.
Specifically, INGP maintains $\mathbf{L}$ levels hash tables, and each table contains $\mathbf{J}$ feature vectors with dimensionality $\mathbf{F}$. We denote the feature vectors in the hash tables as 
$\mathcal{H}=\left\{{\mathcal{H}}_{i} \mid l \in\{1, \ldots, \mathbf{L}\}\right\}$. 
Each table is independent and stores feature vectors at the vertices of a grid with the resolution of $ N_{l}$, which is chosen evenly between the coarsest and finest resolution $N_{\min }$,$N_{\max }$.
Practically, we set $N_{\min }=16$ and $N_{\max }$ is the same as the original resolution of the input videos.\par
We denote the multi-resolution hash encoding with learnable feature vectors $\mathcal{H}$ as $\mathbf{h}(\cdot|\mathcal{H})$.
For a specific level $l$, a 4d-vector $\mathbf{z} \in \mathbf{R}^{4}$ is scaled by that level's resolution and then spans a voxel by rounding up and down$ \left\lceil\mathbf{z}_{l}\right\rceil:=\left\lceil\mathbf{z} \cdot N_{l}\right\rceil$, $\left\lfloor\mathbf{z}_{l}\right\rfloor:=\left\lfloor\mathbf{z} \cdot N_{l}\right\rfloor$. 
The feature of $\mathbf{z}$ is four linear interpolated by the feature vectors at each corner of the voxel. The feature vectors at each corner are queried from $\mathcal{H}_{l}$ using the following spatial hash function\cite{teschner2003optimized}: 
\begin{equation}
    hash(\mathbf{z})=\left(\bigoplus_{i=1}^{4} z_{i} \pi_{i}\right) \bmod \mathbf{T},
\end{equation}
where $\bigoplus$ denotes the XOR operation and $\pi_{i}$ are unique, large prime numbers.
The feature vectors of $\mathbf{z}$ at $\mathbf{L}$ levels are then concatenated to produce $\mathbf{h}(\mathbf{z}|\mathcal{H}) \in \mathbf{R}^{\mathbf{L}\mathbf{F}}$.

\subsection{Model}
\label{representation}
Although INGP\cite{muller2022instant} is capable of converging in a short time and rendering high-fidelity novel view images, it is only suitable for static scenes and needs dense multi-view inputs. 
To aggregate the corresponding point information of different frames, we introduce a surface-relative representation relative to the surface point cloud of the human body (in practice, SMPL\cite{loper2015smpl} or the mesh obtained by SelfRecon\cite{jiang2022selfrecon}). 
We also adopt a multi-resolution hash encoding\cite{muller2022instant} to speed up training. Specifically, for each frame $t$, we maintain the human surface $\mathbf{T}_{t}$, which is used to calculate the surface-relative hash encoded feature vector (see Sec\ref{relative}) $Rel(\mathbf{x}|\mathbf{T}_{t})$ of any given sample point $\mathbf{x}$ in this frame.
We then feed $Rel(\mathbf{x}|\mathbf{T}_{t})$ into MLP $F_\omega$ to predict the radiance and density of that point. Following NeuralBody\cite{peng2021neural}, an optimizable latent embedding $\mathbf{\ell}_t$ for each frame $t$ is employed to encode the temporally-varying factors.
The density and radiance field at frame $t$ can be defined as:
\begin{equation}
    \left(\sigma_{t}(\mathbf{x}), \mathbf{c}_{t}(\mathbf{x})\right) = F_{\omega}\left(Rel(\mathbf{x}|\mathbf{T}_{t})),\mathbf{\ell}_t\right),
\end{equation}
where $\sigma_{t}(\mathbf{x})$ and $\mathbf{c}_{t}(\mathbf{x})$ are the density and radiance of the sample point $\mathbf{x}$ at frame $t$, $F_{\omega}$ represents the MLP function with learnable weights ${\omega}$.

Finally, the color of each sample ray is calculated with Eq.~\ref{volume_rendering}. The photometric loss and geometric guidance loss are used to optimize ${\omega}$, latent embedding $\left\{\ell_{\iota}\right\}_{t=0}^{N_{l}-1}$ and the features $\mathcal{H}$ in hash table.

\begin{equation}
    {\left\{\ell_{\iota}\right\}_{t=0}^{N_{l}-1}, \mathcal{H}, \omega}=\textbf{argmin}\left(L_{\mathrm{rgb}}+L_{\mathrm{geo}}\right),
\end{equation}

where $L_{\mathrm{rgb}}$ and $L_{\mathrm{geo}}$ are the photometric and geometry guidance loss function explained in Sec\ref{loss}.

\subsection{Surface-Relative Hash Encoding}
\label{relative}

\paragraph{Surface-Relative Representation}
Given a sample point $\mathbf{x}$ at frame $t$, we need to construct a relative representation conditioned on the human surface point cloud. This representation aims to ensure that the corresponding points in different frames will be mapped to the same representation and thus get the same feature vector fed into the MLP $F_\omega$ to regress color and density.
Specifically, given two spatial points $\mathbf{x}$ and $\mathbf{x}^{'}$ at different frame $t$ and $t'$ respectively, we should have $Rel(\mathbf{x}|\mathbf{T}_t)=Rel(\mathbf{x}'|\mathbf{T}_{t'})$ if they are in correspondence.
We observe that when the human body moves over time, the k-nearest points of the query point on the surface point cloud and the corresponding signed distances are roughly unchanged. 
Based on this observation, we utilize the k-nearest neighbor vertices and their corresponding signed distance to represent the query point. The reason why we do not use the nearest point on the face like H-NeRF's method\cite{xu2021h} is that the computational costs are relatively high to compute an exact nearest point on the face rather than the vertices, and their representation based on the single closest point leads to artifacts around body joints (e.g., armpits) for unseen poses. To increase the inference speed and solve these artifacts, we use k-nearest vertices on the surface point cloud instead.

In practice, we first calculate the k-nearest points $N_{k}(\mathbf{x})$ of query point $\mathbf{x}$ in the surface point cloud $\mathbf{T}_t$ and the corresponding signed distance value $d_{i}(\mathbf{x})$ through the KNN algorithm. Then we use the k corresponding points of $N_{k}(\mathbf{x})$ in template surface $\mathbf{T}_{0}$ and k signed distances to represent $\mathbf{x}$. Formally, given any point $\mathbf{x}$ in the posed space at frame $t$, we calculate its relative representation as:

\begin{equation}
\begin{array}{c}
\mathcal{N}(\mathbf{x}|\mathbf{T}_{t})=\left\{\mathbf{r}_{1},\mathbf{r}_{2},\cdots,\mathbf{r}_{k}\right\}, \\
\\
\mathbf{r}_{i} = (\bar{\mathbf{p}}_{i},d_{i}(\mathbf{x})),
\end{array}
\end{equation}

\begin{figure}
    \centering

    \includegraphics[width=\linewidth]{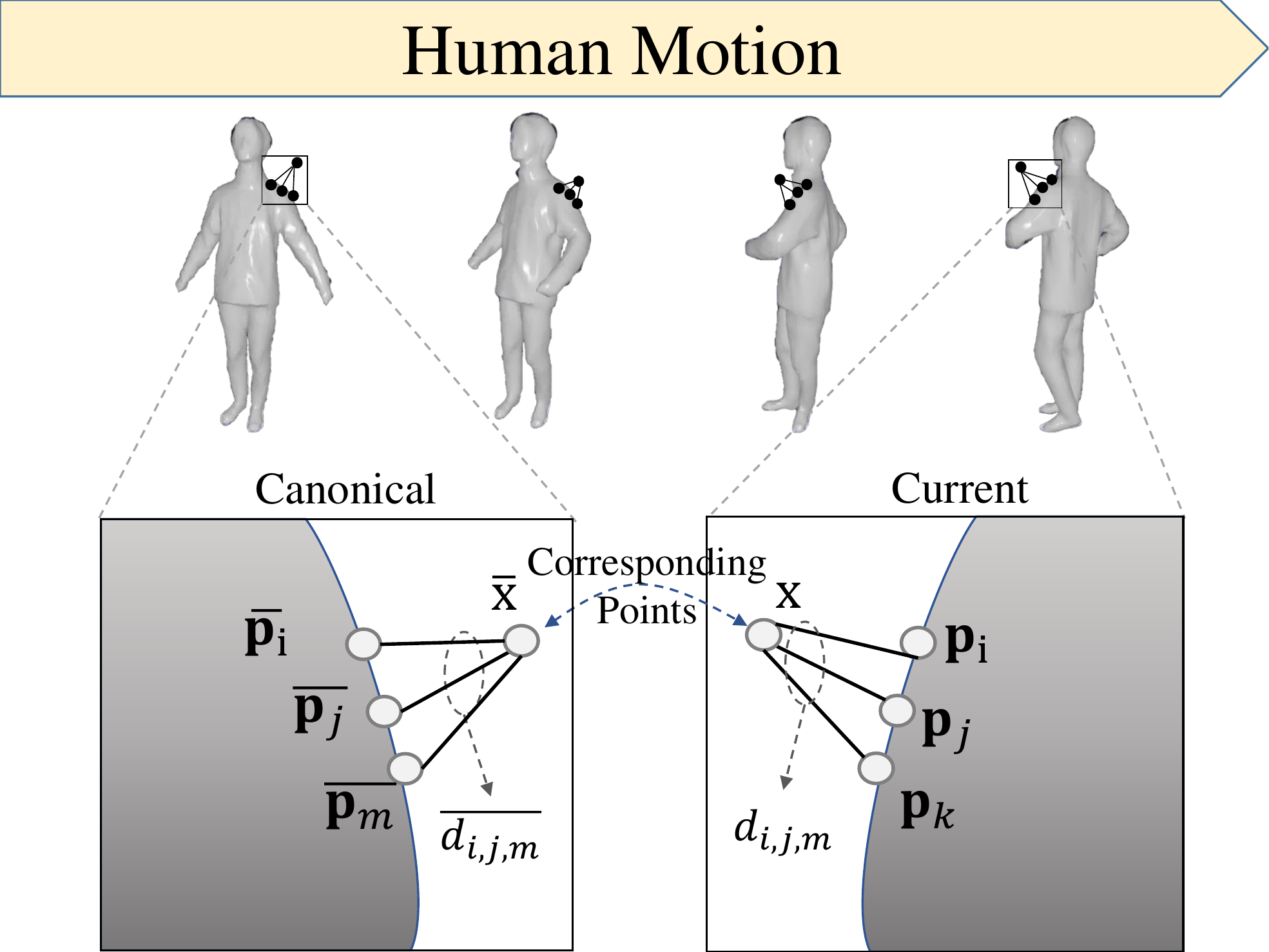}
 
    \vspace*{-2mm}
    \caption{Illustration of our surface-relative representation. The k-nearest vertices $\mathbf{p}_{i,j,k}$ on the human surface of corresponding points stay unchanged when human moves.}
    \label{fig:dataset}
    \vspace*{-5.5mm}
\end{figure}

 where $\mathbf{p}_{i} \in N_{k}(\mathbf{x})$ refers to the coordinate of i-th nearest points of $\mathbf{x}$ in the current surface $\mathbf{T}_t$,  $\bar{\mathbf{p}}_{i}$ means the corresponding point of $\mathbf{p}_{i}$ in $\mathbf{T}_0$ and $d_{i}(\mathbf{x}) \in \mathbf{R}$ is signed distance value from $\mathbf{p}_{i}$ to $\mathbf{x}$. After that, we feed it into the hash encoder and blend the feature vectors:
 \begin{equation}
     Rel(\mathbf{x}|\mathbf{T}_{t})=\frac{\sum w_{i} \mathbf{h} \left( \hat{\mathbf{r}}_{i}|\mathcal{H}\right)}{\sum w_{i}},
 \end{equation}
where $\hat{\mathbf{r}}_{i}$ is normalization of $\mathbf{r}_{i}$, $\mathbf{h}(\cdot|\mathcal{H})$ refers to multi-resolution hash encoding, $w_{i}$ is the blending weight defined as:
\begin{equation}
    w_{i}=|\cos \left(\mathbf{x}-\mathbf{p}_{i}, \mathbf{n}_{i}\right)|,
\end{equation}
and $\mathbf{n}_{i}$ is the normal vector of $\mathbf{p}_{i}$ on the current point cloud $\mathbf{T}_t$. The blended feature vector $Rel(\mathbf{x}|\mathbf{T}_{t})$ is then sent to $\mathcal{F}_\omega$ to compute the color and density of $\mathbf{x}$ at frame $t$.

\subsection{Training}
\label{loss}
We use the following loss function to jointly optimize  ${\omega, \mathcal{H}}$ and latent embedding $\ell_{i}$. 

\subsubsection{Photometric Loss}
We minimize the render error of all observed images, and the loss function is defined as:
\begin{equation}
L_{\mathrm{rgb}}=\sum_{\mathbf{r} \in \mathcal{R}}\left\|\tilde{\mathbf{C}}(\mathbf{r})-\mathbf{C}(\mathbf{r})\right\|_{2},
\end{equation}
where $\mathcal{R}$ is the set of rays passing through image pixels and $\tilde{\mathbf{C}}(\mathbf{r})$ is the ground truth color.

\subsubsection{Geometry Guidance loss}To guide the density field to converge in the direction of the humanoid during the early stage of training, we introduce the geometry guidance loss, which consists of the following two functions:
\paragraph{Mask Loss} Since the human body is non-transparent, if a ray is sampled from the pixel in the masked area, the weight sum should be close to 1; otherwise, it should be close to 0. Thus we require the weight sum of the ray to match the input masks.
\begin{equation}
L_{\mathrm{mask}}=\sum_{\mathbf{r} \in \mathcal{R}}\left ({\mathbf{W}}(\mathbf{r})(1-\mathbf{M}(\mathbf{r}))+(1-\mathbf{W}(\mathbf{r}))\mathbf{M}(\mathbf{r})\right ),
\end{equation}
where $\mathbf{W}(\mathbf{r}) = \sum_{i = 1}^{N} \alpha\left(\mathbf{x}_{i}\right) \prod_{j<i}\left(1-\alpha\left(\mathbf{x}_{j}\right)\right)$ is the weight sum of the ray $\mathbf{r}$, $\mathbf{M}(\mathbf{r})=1$ if $\mathbf{r}$ is sampled from the pixel in the masked area otherwise 0.
 
\paragraph{Distance Loss} If a point is far from the human body, its density should be close to 0. Therefore, an exponential function is used to penalize for density outside the human body.
\begin{equation}
L_{\mathrm{dist}}=\sum_{\mathbf{x} \in \mathcal{X}} \sigma(\mathbf{x}) \exp(\varphi(d_{1}(\mathbf{x}))\beta),
\end{equation}
where $\mathcal{X}$ is the set of points sampled on the rays in $\mathcal{R}$, $\beta$ is a hyper parameter and $\varphi(\cdot)$ refers to the Relu function.
The final geometry guidance loss is defined as:
\begin{equation}
    L_{\mathrm{geo}}=\lambda_{\mathrm{mask}}L_{\mathrm{mask}}+\lambda_{\mathrm{dist}} L_{\mathrm{dist}}
\end{equation}

\subsubsection{Training Strategy}
The total loss function can be represented as:
\begin{equation}
     L_{\mathrm{total}} = L_{\mathrm{rgb}} + \lambda L_{\mathrm{geo}}
\end{equation}
We set $\lambda=1$ in the first 400 iterations to learn a rough geometry of human body, then set $\lambda=0.1$ to learn the geometric details and colors mainly through inverse rendering.

\paragraph{Sample Space Annealing}Following RegNeRF\cite{niemeyer2021regnerf}, we apply sample space annealing to avoid high-density values at ray origins. In practice, we start at a small range around the middle of the ray and gradually increase the sample range as training progresses.

\subsection{Implementation Details}
\label{detail}
We implement our code on top of of the torch-ngp\footnote{https://github.com/ashawkey/torch-ngp} codebase\cite{tiny-cuda-nn,muller2022instant,torch-ngp,tang2022compressible}.
We optimize with Adam\cite{kingma2014adam} using an learning rate decay from $2 \cdot 10^{-3}$ to $2 \cdot 10^{-5}$. 
To speed up evaluation, we render a rough mask of the human performer with pyrender and only apply volume rendering in the masked area. 
For a 540$\times$540 monocular video of 200 frames, we need around 3K iterations to converge (about 12 minutes on a single NVIDIA GeForce GTX3090 GPU).

\section{Experiments}

To demonstrate the effectiveness of our method, we conduct comparison experiments on monocular videos. Some ablation studies are also discussed to evaluate the necessity of our modules.
\subsection{Dataset}
\label{data}
To evaluate our method's reconstruction ability from single-view input, we capture some custom data, which includes monocular videos of human performers and the corresponding high-fidelity template meshes reconstructed by SelfRecon\cite{jiang2022selfrecon}.
Each video contains the whole body information of the performer from a single view. Videos from other views are provided for evaluation.\par

We also employ ZJU-Mocap\cite{peng2021neural} dataset
for comparing with state of the art methods. 
For each object in these datasets, we adopt SelfRecon\cite{jiang2022selfrecon} to reconstruct the high-fidelity surface to better aggregate the information across frames. Some generated models, such as ICON\cite{xiu2021icon} can do the same much faster but with less precision.
\paragraph{metrics} Consistent with NeRF\cite{mildenhall2020nerf}, we use two standard metrics to quantify novel view synthesis results: peak signal-to-noise ratio (PSNR) and structural similarity index (SSIM). 
Following NeuralBody\cite{peng2021neural}, we only calculate these metrics on pixels inside the 2D bounding box, which is obtained per view from the input masks.

\subsection{Evaluation}



We compare with state-of-the-art implicit human novel view synthesis methods that based on NeRF:\par 

1) AnimatableNeRF\cite{peng2022animatable} utilizes deformation fields based on neural blend weight fields to aggregate per-frame information to reconstruct the canonical space's neural radiance field.\par

2) NeuralBody\cite{peng2021neural} reconstructs per frame's neural radiance field conditioned at body structured latent codes, which are diffused to the whole space by the SparseConvNet.\par
We perform this experiment on
ZJU-Mocap\cite{peng2021neural} and our custom data. 
Specifically,
we choose 4 objects (313, 315, 377, 386) with relatively high image quality and use "Camera (1)" for training and other views for evaluation.

We use the official open-source code of both NeuralBody and AnimatableNeRF to compare with our method. For fair comparisons, we have trained NeuralBody and AnimatableNeRF from scratch for around twenty hours with the same data and use SelfRecon meshes instead of SMPL in NeuralBody.\par

\begin{figure*}
    \centering

     \includegraphics[width=\linewidth]{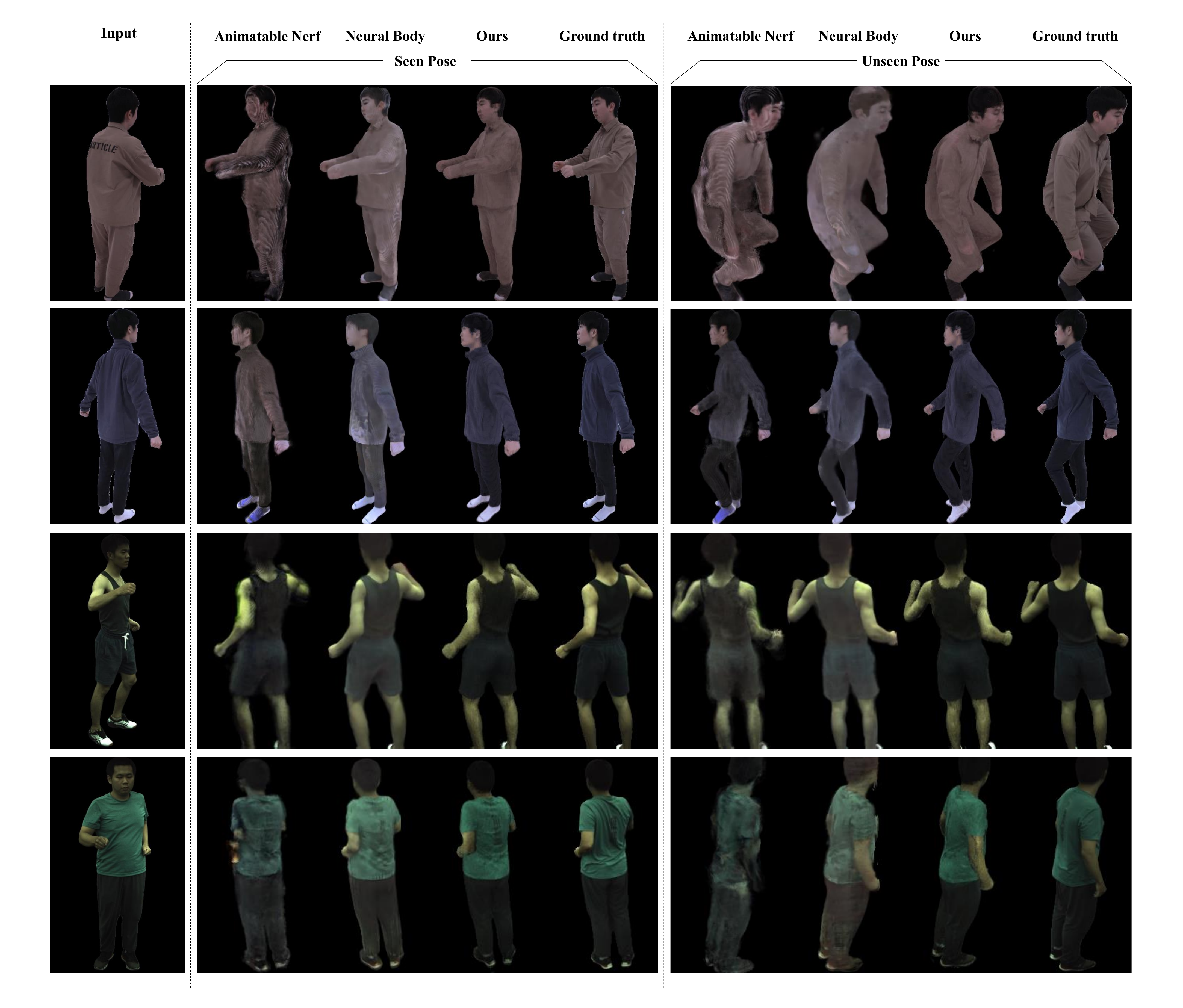}

    \vspace*{-4mm}
    \caption{Qualitative comparison result on ZJU-MoCap dataset and our custom data}
    \label{fig:compare}
    \vspace*{-4mm}
\end{figure*}
\begin{figure*}[thb]
    \centering

    \includegraphics[width=\linewidth]{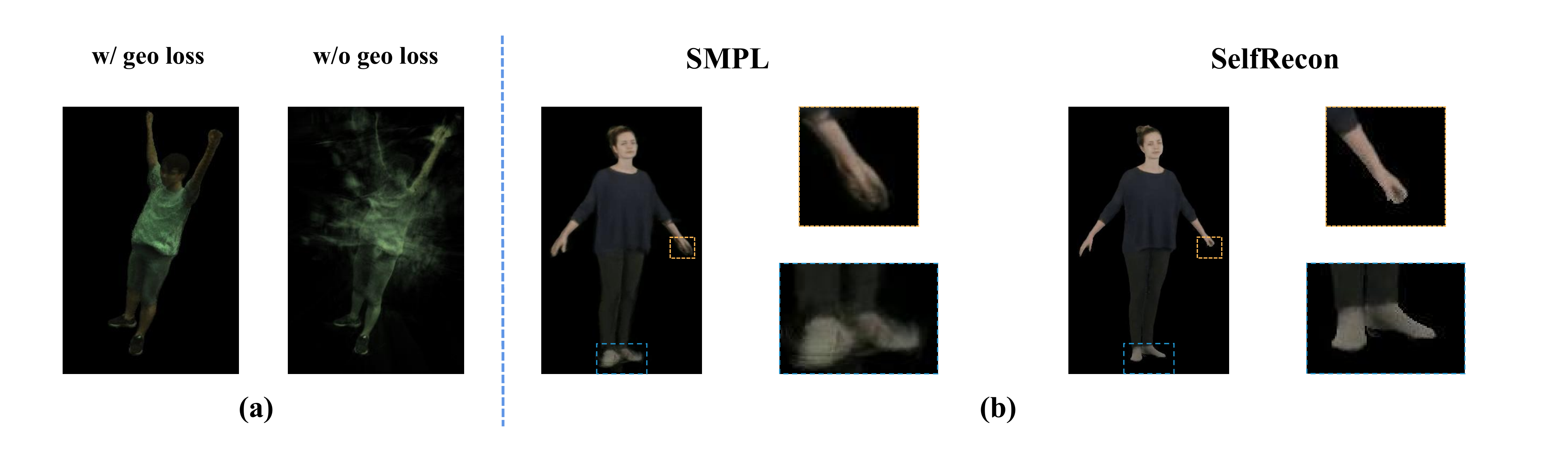}

    \vspace*{-3mm}
    \caption{(a) Ablation study on geometry guidance loss. The geometry guidance loss guides the model to learn the correct geometric surface.
    (b) Ablation study on human surface. With a more accurate human surface, our method can generate results with more details.}
    \label{fig:ablation}
    \vspace*{-3mm}
\end{figure*}
As shown in Fig.~\ref{fig:compare}, our method can produce satisfactory results similar to the ground truth even on completely unobserved views, while NeuralBody and AnimatableNeRF tend to produce results with blurs and color differences. 
Fig.~\ref{fig:speed} illustrates the coverage speed, our method is able to achieve reasonable results for a 200-frames monocular video in twenty minutes. Tab.~\ref{tab:1} and Tab.~\ref{tab:2} shows the quantitative results for training frames and novel poses respectively, our method outperforms NeuralBody and AnimatableNeRF for all datasets and under all metrics. We observed that the results of all methods on ZJUMocap datasets in the unseen pose case are better than those in the seen pose case. This is caused by the small proportion of the human bodies in the 2D bounding box during the unseen pose sequence we selected.
\begin{table}[t]
    \begin{center}
    \resizebox{\linewidth}{!}{
    \begin{tabular}{|l|c|c|c|c|c|c|c|c|}
    \hline
                  & \multicolumn{2}{c|}{ZJUMocap} & \multicolumn{2}{c|}{Custom Data} \\
    \hline
    Method        & PSNR $\uparrow$ & SSIM $\uparrow$   & PSNR $\uparrow$ & SSIM $\uparrow$  \\
    \hline\hline
    NeuralBody~\cite{peng2021neural}&23.47&0.862&22.23&0.847\\
    AnimatableNeRF~\cite{peng2022animatable}       &23.48&0.857&23.67&0.833\\
    Ours          &\textbf{24.85}&\textbf{0.875}&\textbf{26.02}&\textbf{0.880}\\
    \hline
    \end{tabular}}
    \end{center}
    \vspace{-5mm}
    \caption{Quantitative comparison for training frames.}
    \label{tab:1}
    \vspace{-3mm}
\end{table}

\begin{table}[t]
    \begin{center}
    \resizebox{\linewidth}{!}{
    \begin{tabular}{|l|c|c|c|c|c|c|c|c|}
    \hline
                  & \multicolumn{2}{c|}{ZJUMocap} & \multicolumn{2}{c|}{Custom Data} \\
    \hline
    Method        & PSNR $\uparrow$ & SSIM $\uparrow$   & PSNR $\uparrow$ & SSIM $\uparrow$  \\
    \hline\hline
    NeuralBody~\cite{peng2021neural}&23.62&0.864&21.99&0.842\\
    AnimatableNeRF~\cite{peng2022animatable}       &23.81&0.857&22.50&0.807\\
    Ours          &\textbf{25.37}&\textbf{0.870}&\textbf{25.49}&\textbf{0.873}\\
    \hline
    \end{tabular}}
    \end{center}
    \vspace{-5mm}
    \caption{Quantitative comparison for novel poses.}
    \label{tab:2}
    \vspace{-7mm}
\end{table}

\subsection{Ablation Study}

\begin{figure}
    \centering

    \includegraphics[width=\linewidth]{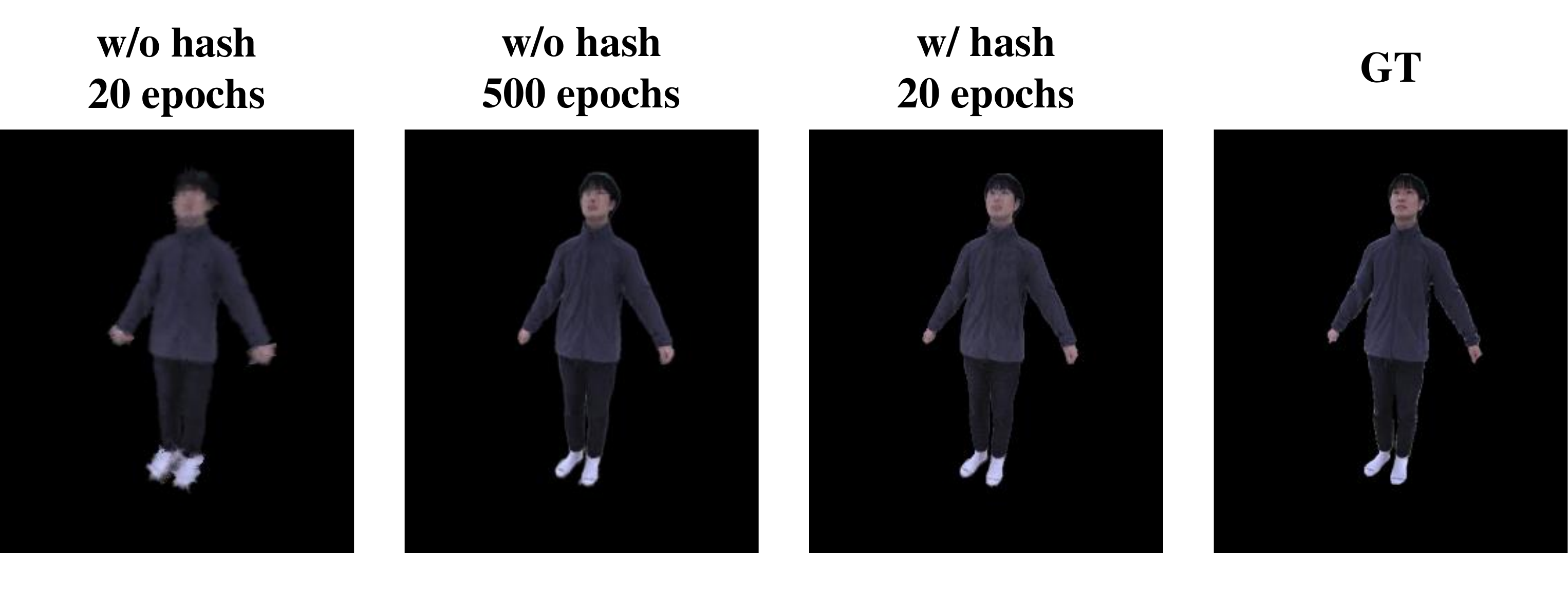}

    \vspace*{-3mm}
    \caption{Ablation study on hash encoding.}
    \label{fig:ablation_hash}
    \vspace*{-3mm}
\end{figure}

\paragraph{Ablation Study on Hash Encoding} 
To further verify the effectiveness of hash encoding in our method, we design the following baseline version. All the feature vectors originally obtained by hash encoding are now independent and optimizable, and we denote it as vertex-based representation. In practice, for each vertex in the template mesh, we anchor 500($\approx N_{max}$) optimizable feature vectors with dimensionality 32 and directly optimize these feature vectors during training. Then we use our custom data to train the origin model and vertex-based representation, respectively.
The corresponding rendering results are shown in Fig.\ref{fig:ablation_hash}.
The vertex-based representation takes at least ten times longer to converge. Using multi-resolution hash encoding dramatically increases the training speed of our method.

\paragraph{Ablation Study on the Geometry Guidance Loss} We attempt to remove the geometry guidance loss from the total loss. As shown in Fig.~\ref{fig:ablation}(a), the geometry guidance loss does guide the model to converge in the direction of the humanoid, preventing it from converging to other local optimal results.
\paragraph{Ablation Study on Human Surface} We discuss the choice of the human surface (SMPL or SelfRecon) in our algorithm. For a monocular video in the People-Snapshot dataset, we use both SMPL and SelfRecon meshes as the human surface and compare their rendering results.
As shown in Fig.~\ref{fig:ablation}(b), with a more stable human pose estimation and non-rigid deformation, the human surface obtained by SelfRecon enables our model to find a better correspondence across frames and thus could generate results with better details.

We did an additional experiment with monocular video inputs to evaluate our method with more accurate SMPL meshes to show how these two points influence the final result. We use multi-view inputs in EasyMocap to obtain more accurate SMPL parameters and one view to train the model. In this experiment, we employ the ZJU-Mocap dataset instead of the People-Snapshot dataset since multi-view inputs are needed to get more accurate SMPL parameters.


 
\begin{figure}[H]
    \centering

    \includegraphics[width=\linewidth]{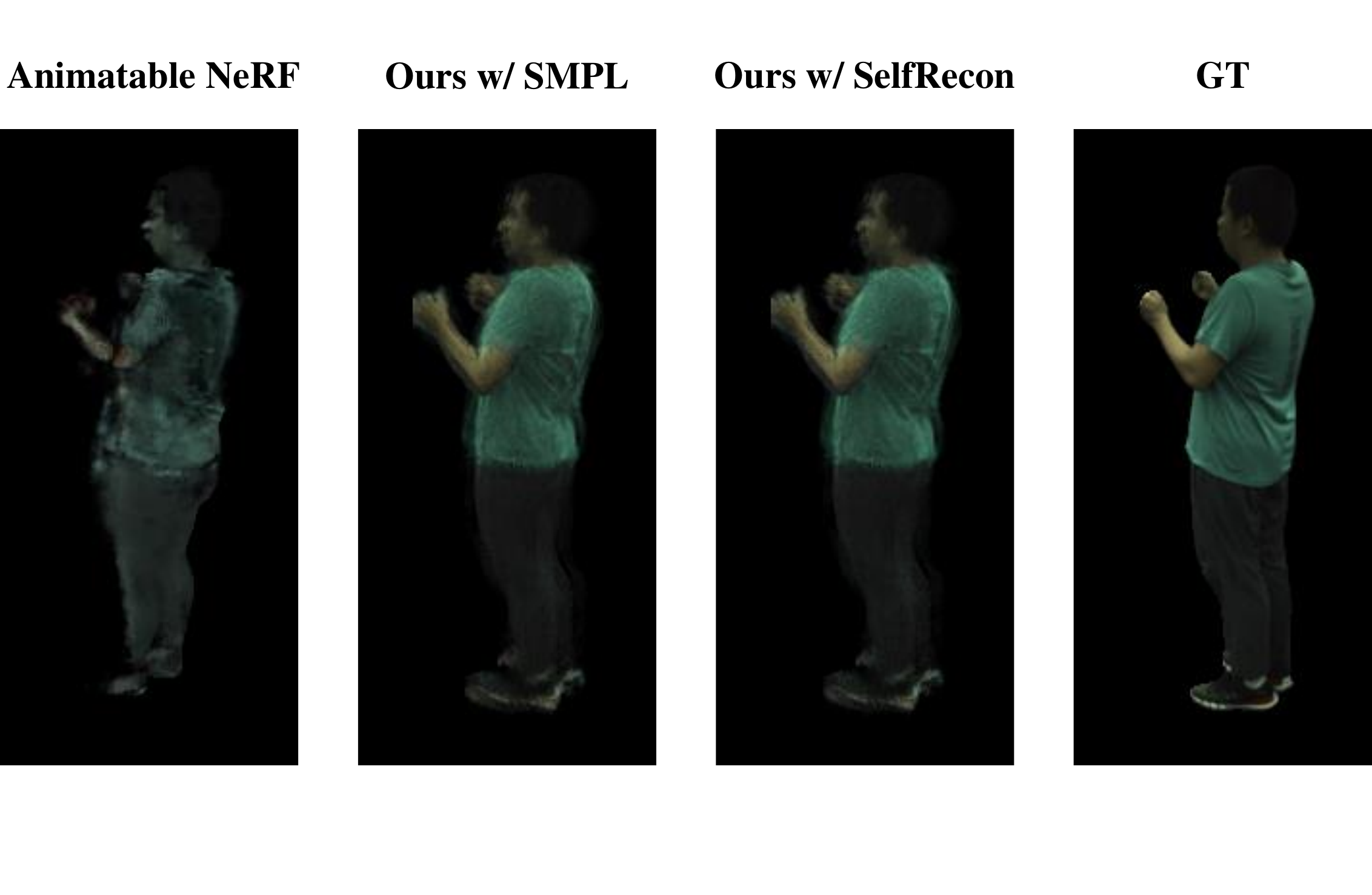}

    \vspace*{-10mm}
    \caption{ablation study on human surfaces.}
    \label{fig:ablation_surface}
    \vspace*{-6mm}
\end{figure}

As shown in Fig.\ref{fig:ablation_surface} and Tab.\ref{tab:3} , our method with accurate SMPL meshes still outperforms NeuralBody\cite{peng2021neural} and AnimatableNeRF\cite{peng2022animatable}. This result demonstrates that if the SMPL parameters are accurate enough, our method can also achieve high-quality results without SelfRecon meshes in most cases.

\begin{table}[t]
    \begin{center}
    \resizebox{\linewidth}{!}{
    \begin{tabular}{|l|c|c|c|c|c|c|}

    \hline
    Method        & PSNR $\uparrow$ & SSIM $\uparrow$    \\
    \hline\hline
    AnimatableNeRF~\cite{peng2022animatable} &23.48&0.857\\
    Ours with accurate SMPL          &24.69&\textbf0.871\\
    Ours with SelfRecon     &\textbf{24.85}&\textbf{0.875}\\
    \hline
    \end{tabular}}
    \end{center}
    \vspace{-2mm}
    \caption{Quantitative ablation study on human surfaces.}
    \label{tab:3}
    \vspace{5mm}
\end{table}

\section{Limitation and Future Work}
\label{limitation}
Currently, SelfNeRF assume that some details like human fingers and wrinkles of the clothes move consistent with the input human surface sequence; thus, blurring might occur when the input videos have frequently changed details or input the human surface sequence is not accurate and stable enough. This problem might be solved by adding a learnable per-frame offset, and we leave it as our future work.

Second, although using SMPL meshes with accurate pose parameters as human surface are sufficient in most cases, we still need a more accurate human surface in some cases. Moreover, our data preparation takes a long to process if we require a more accurate human surface as input, such as SelfRecon. However, we believe that this problem will be solved as the accuracy and speed of human shape reconstruction methods have already been greatly improved\cite{xiu2021icon,alldieck2022photorealistic}.

\section{Conclusion}
We have proposed SelfNeRF, an efficient novel view synthesis method for dynamic human bodies from monocular self-rotating inputs based on neural radiance field. We introduced a novel surface relative representation based on the KNN algorithm, which could aggregate information across time and extend the multi-resolution hash encoding in INGP from static scene to dynamic human shapes. In this way, SelfNeRF only requires monocular self-rotating inputs and could converge to high-quality  result with about twenty minutes. Extensive experimental results have showed that we can generate high-fidelity results for this challenging task, demonstrating potential practical applications of SelfNeRF.
\paragraph{Acknowledgements}
This research was supported by the National Natural Science Foundation of China (No.62122071, No.62272433), and the Fundamental Research Funds for the Central Universities (No. WK3470000021).

{\small
\bibliographystyle{ieee_fullname}
\bibliography{egbib}
}

\end{document}